%% file: dignet_paper.tex
\documentclass[twoside,11pt]{article}
\usepackage{jair, theapa, rawfonts}
\usepackage[misc]{ifsym}
\usepackage{ulem}
\usepackage{soul}
\usepackage{url}
\usepackage{latexsym}
\usepackage[utf8]{inputenc}
\usepackage{multicol}
\usepackage{caption}
\usepackage{multirow}
\usepackage{graphicx}
\usepackage{amsmath}
\usepackage{booktabs}
\usepackage{algorithm}
\usepackage{algorithmic}
\usepackage{stfloats}
\usepackage{comment}
\usepackage{ulem}
\usepackage{color}
\usepackage{array}
\usepackage{booktabs}
\usepackage{amsfonts}

\ShortHeadings{DigNet: Digging Clues from Local-Global Interactive Graph for ASC}
{Xing, Tsang}

\begin{document}

%
\title{DigNet: Digging Clues from Local-Global Interactive Graph for Aspect-level Sentiment Classification}

\author{\name Bowen Xing \email Bowen.Xing@student.uts.edu.au \\
       \addr Australian Artificial Intelligence Institute, University of Technology Sydney \\Ultimo, NSW 2007, Australia
       \AND
       \name Ivor W. Tsang \email ivor\_tsang@ihpc.a-star.edu.sg \\
       \addr Centre for Frontier AI Research, A*STAR, 138632, Singapore\\ Australian Artificial Intelligence Institute, University of Technology Sydney\\ Ultimo, NSW 2007, Australia
       }
\maketitle

\begin{abstract}
In aspect-level sentiment classification (ASC), state-of-the-art models encode either syntax graph or relation graph to capture the local syntactic information or global relational information.
Despite the advantages of syntax and relation graphs, they have respective shortages which are neglected, limiting the representation power in the graph modeling process.
To resolve their limitations, we design a novel local-global interactive graph, which marries their advantages by stitching the two graphs via interactive edges.
To model this local-global interactive graph, we propose a novel neural network termed DigNet, whose core module is the stacked local-global interactive (LGI) layers performing two processes: intra-graph message passing and cross-graph message passing.
In this way, the local syntactic and global relational information can be reconciled as a whole in understanding the aspect-level sentiment.
Concretely, we design two variants of local-global interactive graphs with different kinds of interactive edges and three variants of LGI layers.
We conduct experiments on several public benchmark datasets and the results show that we outperform previous best scores by 3\%, 2.32\%, and 6.33\% in terms of Macro-F1 on Lap14, Res14, and Res15 datasets, respectively, confirming the effectiveness and superiority of the proposed local-global interactive graph and DigNet.
\end{abstract}


\input{introduction}

\input{relatedwork}

\input{graphconstruction}
\input{framework}

\input{experiment}

\input{conclusion}

\section*{Acknowledgements}
This work was supported by Australian Research Council  Grant (DP180100106 and DP200101328).
Ivor W. Tsang was also supported by A$^*$STAR Centre for Frontier AI Research (CFAR).

\vskip 0.2in
\bibliography{ref}
\bibliographystyle{theapa}

\end{document}

%% file: introduction.tex
\section{Introduction}\label{sec:introduction}
Aspect-level sentiment classification (ASC) \cite{TDLSTM,ATAE} is a subtask of aspect-based sentiment analysis \cite{tkdesurvey}. ASC aims to predict the sentiment polarity of an aspect mentioned in a review context. And the aspect is a subsequence in the context word sequence. For example, suppose that a review is ``I like the noodles here, although the service is bad.'' and the given aspects are `noodles' and `service', whose sentiment polarities are positive and negative, respectively.

\begin{figure*}[t]
 \centering
 \includegraphics[width = 1.0\textwidth]{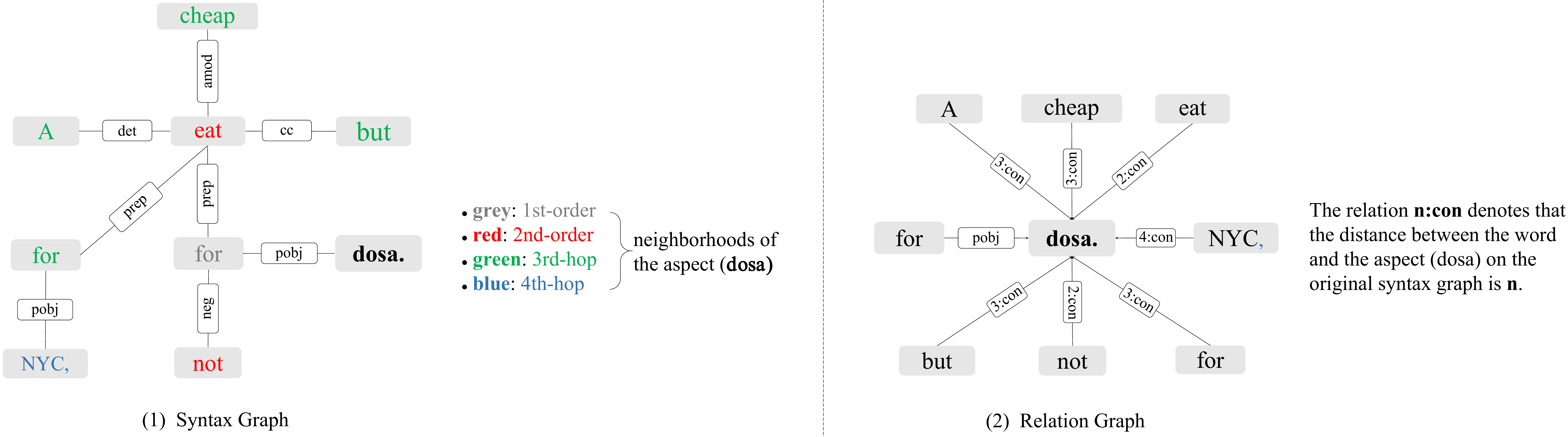}
 \caption{Syntax graph and relation graph of the sample ``A cheap eat for NYC, but not for dosa.'' and the aspect is `dosa'.}
 \label{fig: dual-view}
\end{figure*}

Recently, syntax graphs output by off-the-shelf dependency parsers have been widely used and shown their advantages in the ASC task.
Encoding the syntactic structure of the syntax graph can capture better correlations between the aspect and its relevant words depicting its sentiment \cite{asgcn,yuezhangRGAT}.
An example of a syntax graph is shown in Fig. \ref{fig: dual-view} (1). 
A stream of works \cite{asgcn,graphatt,DGEDT,bigcn} employ graph convolutional networks (GCN \cite{gcn}) and graph attention network (GAT \cite{gat}) to encode the context regarding the adjacent matrix derived from the structure of  a syntax graph.
A key advantage of such syntax graph is that as its message passing is based on local syntactic connections, so the local clues for ASC can be easily captured.
For example, aspect-related words may be far from the aspect in a context word sequence, but they are the aspect's local neighborhoods on the syntax graph.
In this case, GCN/GAT can syntactically draw these relevant words to the aspect, while only relying on attention mechanisms may lose the aspect-related information contained in these words \cite{asgcn}.


More recently, \cite{RGAT} propose a relation graph and a relational graph attention network (R-GAT) which operates on the relation graph and aggregates the global relational information from each context word to the aspect node.
An example of a relation graph is shown in Fig. \ref{fig: dual-view} (2).
It is obtained by reshaping and pruning the original syntax graph to directly connect each context word to the aspect with relation labels which are defined regarding original dependency labels.
The key advantage of the relation graph is that its message passing is aspect-aware due to the aspect-centric structure, and the message passing is only 1-hop, which is efficient without message vanishing.

Despite the advantages of the two graphs, we find that they have their respective shortages, which are neglected by previous works.
As for the syntax graph, its local-connection structure not only brings advantages but also causes several issues.
First, the message passing on the syntax graph is aspect-agnostic since GCN and GAT do not know which node is the aspect node.
Second, if the aspect-relevant nodes are not local neighborhoods of the aspect node on the syntax graph, they cannot exchange information with the aspect node.
The reason is that $n-$layer GCN/GAT can only achieve the message passing between a node with its $n-$hop neighborhoods, and $n$ is usually set to 2 in previous works \cite{asgcn,CDT,bigcn}.
For example, in Fig. \ref{fig: dual-view} (1), `cheap' and `but' are two important words that express the sentiment of `dosa', while their distances to `dosa' are 3.
In this case, a 2-layer GCN/GAT cannot achieve the message passing between these two words and `dosa', harming aspect sentiment reasoning.
This issue cannot be resolved by increasing $n$ because larger $n$ leads to over-fitting,  degrading the performance on test set \cite{asgcn,bigcn}.
Third, the critical information may vanish in the multi-hop message passing process of GCN/GAT.

As for the relation graph, due to its aspect-centric star-shaped structure, there are only star-shaped global connections, while every two non-aspect nodes are isolated.
This isolating issue hinders capturing local clues.
For example, in Fig. \ref{fig: dual-view} (2), it is intuitive that the semantics of `a cheap eat' and `but' should be combined for deducing the negative sentiment of `dosa', but on the relation graph, this cannot be achieved due to the isolating of each two non-aspect words.
Besides this, since all message passings are from non-aspect nodes to the aspect node, non-aspect nodes cannot be updated, which hinders the multi-layer recurrently node updating, which is potential to improve performance.
Previous works did not mediate the shortages of syntax graph and relation graph, and they only leveraged either one of them.
As a result, the captured clues are insufficient and the shortages of the two graphs are inherited, which leads to limited performance gain and has room for improvement.

Based on the above observations, we find that the respective advantages of syntax and relation graphs mediate each other's shortages.
Then this leaves us with a question: \textit{Can we stitch these two graphs and let them compensate for each other via remedying their shortages using each other's advantages?}
Theoretically, we can design some interactive edges to stitch the two graphs.
And the interactive edges can let the two graphs share the beneficial information and supplement each other's missing information.
In this way, the two graphs can compensate for each other through the knowledge transfer along the interactive edges, and it is promising that their respective node representations can be improved (this is proven in Sec. \ref{sec: ablate} RQ4).
Then the integration of the two graphs derives more sufficient information, which combines the advantages of both graphs.

Motivated by this, in this paper, we design a novel local-global interactive graph (LGIG) that marries syntax graph and relation graph using interactive edges, on which the beneficial information can be exchanged between the two graphs.
And different kinds of interactive edges form different variants of LGIG.
In this process, we modify the syntax graph by merging the aspect word nodes into a single aspect node and alleviate the second issue of the relation graph by adding a reversed edge from the aspect node to each non-aspect node.
To encode and model LGIG, we propose a novel model named DigNet, whose core is the stacked multiple local-global interactive (LGI) layers.
Each LGI layer performs two processes: intra-graph message passing (IGMP) and cross-graph message passing (CGMP).
On a syntax graph, a position-weighted GCN (PW-GCN) achieves IGMP.
For the IGMP on relation graph, we propose a reversible relational attention network (R$^2$ATN) to achieve the bidirectional message passing between the aspect node and each non-aspect node.
Different from R-GAT \cite{RGAT} which only learns the representation of the aspect node, our R$^2$ATN can update all nodes and is conductive to multi-layer stacking.
As for CGMP, we propose three variants for cross-graph interactions: cross-graph gate, cross-graph multi-layer perception, and cross-graph multi-head attention.

We believe that CGMP can let syntax graph and relation graph benefit from each other's advantages, which make up for their own shortages.
For the syntax graph, CGMP can provide first-order information containing the global relational information between the aspect node and non-aspect nodes.
Hence each non-aspect node can receive the strong aspect-centric signals, which make syntax graph's nodes representations aspect-aware, then assist PW-GCN aggregate more aspect-related information and discard useless information.
For the relation graph, CGMP can provide local semantics containing some clues which can assist R$^2$ATN to learn more reasonable node representations.
In doing so, both local syntactic information and global relational information can be discovered, leveraged, and combined, promoting digging clues.

To the best of our knowledge, this work makes the first attempt to marry syntax graph and relation graph and let them compensate for each other.
Experiments over several public benchmarks demonstrate the effectiveness and superiority of LGIG and DigNet, showing that DigNet strikes a new state-of-the-art benchmark.
We also perform ablation studies to delve into the contribution of each component in DigNet and elucidate why our method performs well, proving that LGI can indeed improve the node representations on both syntax graph and relation graph.
By investigating the effect of varying GCN layer numbers, we prove that DigNet adopts the same layer number as previous works, so the improvements come from the superior of LGI rather than increasing the distance of message passing on syntax graph.
Moreover, the effect of LGI layer number is also investigated, demonstrating the necessity of multi-layer stacking.

%% file: relatedwork.tex
\section{Related Works}\label{sec: relatedwork}
In recent years, most mainstream ASC models focus on developing deep learning neural networks.
According to model structures, existing models can be divided into two main categories, termed sequence-based methods and graph-based methods.

\textbf{Sequence-based Methods}.
This kind of models adopt flat structures, without utilizing syntax information.
In these models, the most common way to model the aspect-context interactions is attention mechanism \cite{ATAE,IAN,Tencent,songle2019,AA-LSTM}.
\cite{IAN} propose an interactive attention mechanism to allow the aspect and context to select information from each other.
\cite{songle2019} devise a self-supervised attention learning method to enhance the attention mechanism with self-mined supervision information.
Except for attention mechanisms, memory networks (MNs) \cite{DMN,tsmn} and convolutional neural networks (CNNs) \cite{pcnn,gcae} are also employed to extract the aspect-related features from hidden states.

\textbf{Graph-based Methods}.
A number of recent works employ graph neural networks such as graph convolutional netowrks (GCN) \cite{gcn,gcn,CDT,DGEDT,tgcn} and graph attention networks (GAT) \cite{gat,graphatt,sagat} to encode the syntax graph predicted by off-the-shelf dependency parsers.
\cite{asgcn} employ a GCN to capture syntactic features and facilitate the information exchange between the aspect and its related context words.
\cite{DGEDT} propose a transformer-based DGEDT model to jointly leverage the flat textual semantics and dependency graph empowered knowledge.
\cite{RGAT} introduce the relation graph which is aspect-centric and star-shaped.
The relation graph is obtained by reshaping and pruning the original syntax graph and the relations are defined regarding original dependency labels.
They also propose a relational graph attention network (R-GAT) which operates on the relation graph to learn the final aspect representation via aggregating information from non-aspect nodes to the aspect node.
\cite{tgcn} leverages the dependency types in the proposed T-GCN and uses an attentive layer ensemble to learn the comprehensive representation from different T-GCN layers.

Our work is in the line of graph-based methods.
Different from existing graph-based methods, we 1) reveal the neglected shortages of syntax graph and relation graph, and find that their shortages correspond to each other's advantages; 2) marry syntax graph and relation graph in the proposed local-global interactive graph (LGIG); 3) propose a novel DigNet model which operates on LGIG and transfers beneficial information between the two graphs, allowing them to compensate for each other.

%% file: graphconstruction.tex
\section{Graph Construction}
\subsection{Syntax Graph}
\begin{figure}[ht]
 \centering
 \includegraphics[width = 0.8\textwidth]{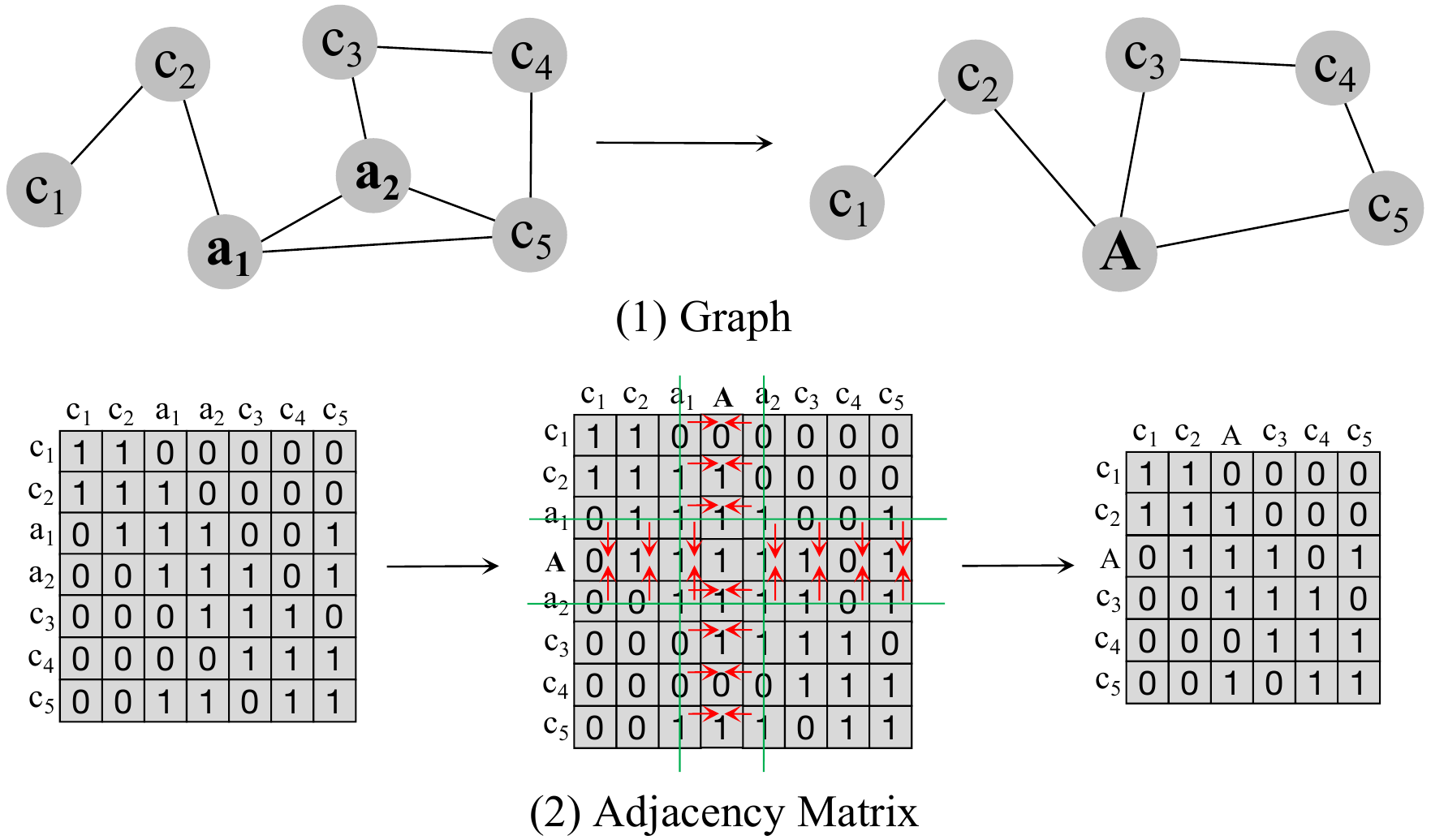}
 \caption{Illustrations of how we merge the multi-word aspect into a single node. Green arrows denote the process of Eq.[\ref{eq:x1},\ref{eq:x2}]. Red arrows denote columns/rows removing.}
 \label{fig: viewx}
\end{figure}
On the relation graph, the aspect holds a single node.
To make syntax graph and relation graph consistent on nodes, we should merge all aspect word nodes on the original syntax graph into a single aspect node $A$.
Fig. \ref{fig: viewx} (1) illustrates how we reshape the syntax graph.
Firstly, a new aspect node $v_a^x$ is added to the original syntax graph.
Then all of the edges between aspect word nodes and non-aspect context words are merged to $v_a^x$.
Finally, all original aspect word nodes are deleted.
In practice, only the adjacency matrix participates in the computation, so we devise a set of matrix operations to achieve the aspect nodes merging on the adjacency matrix.
This process is illustrated in Fig. \ref{fig: viewx} (2).
Firstly, we add a new column and row for $v_a^x$.
Then the column and row elements of $v_a^x$ is derived as: 
\begin{gather}
m_{\tau,i}=
\begin{cases}
0& {s_{\tau, i}=0}\\
1& {s_{\tau, i}\geq 1}\\
\end{cases},\ s_{\tau, i} = \sum_{k=p(a_1)}^{p(a_{N_a})} m_{k,i} \label{eq:x1} \\
m_{i, \tau}=
\begin{cases}
0& {s_{i, \tau}=0}\\
1& {s_{i, \tau}\geq 1}\\
\end{cases} , \ s_{i, \tau} = \sum_{k=p(a_1)}^{p(a_{N_a})} m_{i, k} \label{eq:x2}
\end{gather}
where $\tau$ is the index if node $v_a^x$, and $p(a_i)$ denotes the index of the $i-$th aspect words.
Finally, the rows and columns corresponding to original aspect words are removed from the adjacent matrix.

\subsection{Relation Graph}
On the relation graph, each aspect is treated as a single node and each non-aspect node is directly connected to the aspect node $v_a^y$ with a defined relation.
If a node is a first-order neighborhood of the aspect on the original syntax graph, the relation between it and $v_a^y$ is the original dependency relation.
Otherwise, the relation is defined as $r^i = n:con$, where $n$ is the distance between node $v_i^y$ and $v_a^y$ on the original syntax graph.
As we state in Sec. \ref{sec:introduction}, the standard relation graph in \cite{RGAT} suffers an issue that as all edges are from non-aspect nodes to $v_a^y$, so the non-aspect nodes cannot be updated, hindering the cross-graph interactions, which is the core of this work.

To solve this issue, we augment the standard relation graph with a reversed edge from $v_a^y$ to each non-aspect node, and this reversed edge is labeled with a reversed relation $r_{rv}^i$.
In this way, there is directly bidirectional communication between $v_a^y$ and each non-aspect node which then can be updated.
Besides, there is a 2-hop connection between every two non-aspect nodes, which alleviates the relation graph's isolating issue and helps capture local clues.

\subsection{Local-Global Interactive Graph}
To marry syntax graph and relation graph, we design a Local-Global Interactive Graph (LGIG) which uses \textit{interactive edges} to connect the local syntactic structure of syntax graph and the global relational structure of relation graph.
Generally, a LGIG can be denoted by $\mathcal{G}=(\mathcal{V,E,R)}$.
The nodes in $\mathcal{V}$ are from either syntax graph or relation graph.
The edge $(i,j,r_{ij})\in \mathcal{E}$ represents the message passing from $v_i$ to $v_j$, where $r_{ij}$ is defined as:
\begin{gather}
r_{ij}=
\begin{cases}
0& {\mathcal{I}(v_i)=\mathcal{I}(v_j)=X}\\
r_{ita}&{\mathcal{I}(v_i)\neq \mathcal{I}(v_j)}\\
r^j_{rv} & {\mathcal{I}(v_i)=\mathcal{I}(v_j)=Y, v_i=v^y_{a}}\\
r^i & {\mathcal{I}(v_i)=\mathcal{I}(v_j)=Y, v_j=v^y_{a}}
\end{cases}
\end{gather}
where $\mathcal{I}(v_i)$ is a function that identifies which graph $v_i$ is from, $X$ and $Y$ denote syntax graph and relation graph respectively, and $r_{ita}$ denotes that the edge is an \textit{interactive edge} which achieves the cross-graph connections.
If there is an interactive edge from $v_i$ to $v_j$, the message of $v_i$ can be passed to $v_j$, but they still belong to different graphs.
The message passing on interactive edges only participate CGMP but not IGMP.
\begin{figure}[t]
 \centering
 \includegraphics[width = 0.8\textwidth]{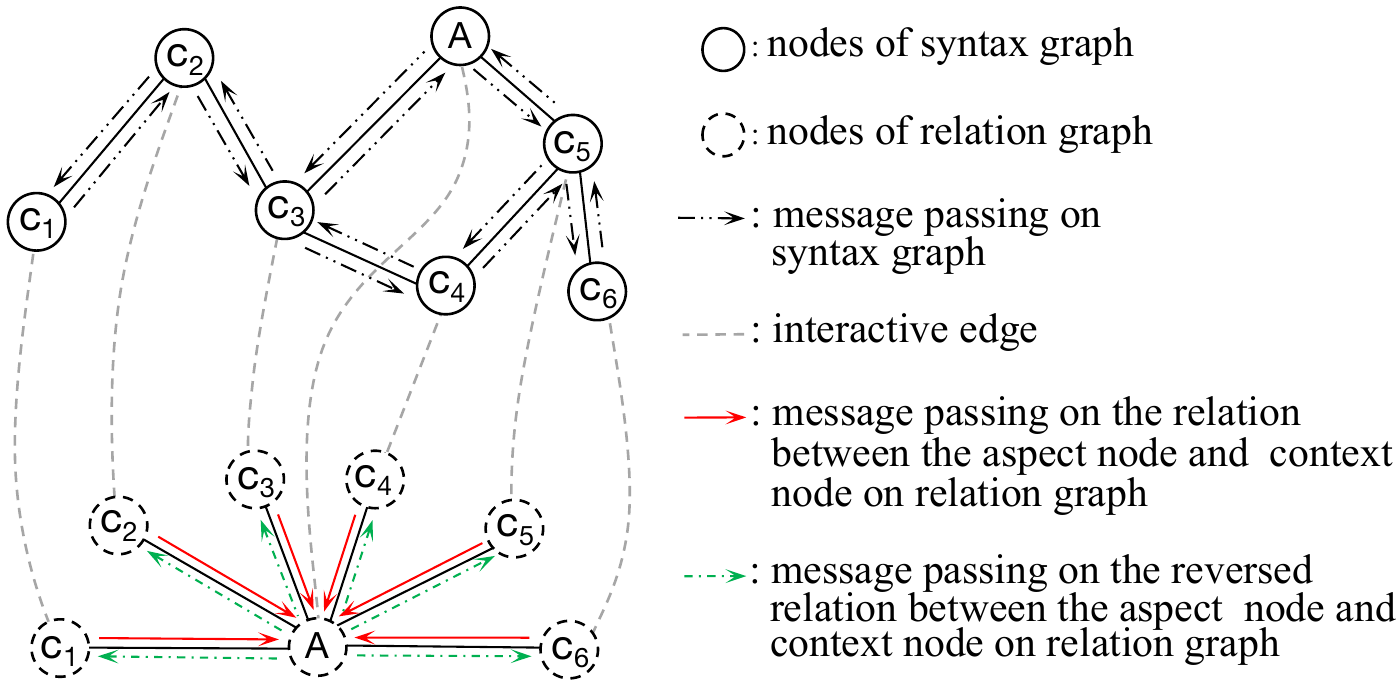}
 \caption{One-to-one local-global interactive graph.}
 \label{fig: dvig}
\end{figure}

We propose two variants of LGIGs with different kinds of cross-graph connections: one-to-one (O2O) LGIG and one-to-all (O2A) LGIG.
An example of O2O-LGIG is shown in Fig. \ref{fig: dvig}, each node is connected to its counterpart node on another graph.
And on O2A-DIG, each node is connected to all nodes on another graph.

%% file: framework.tex

\section{DigNet}\label{sec: framework}
\begin{figure*}[t]
 \centering
 \includegraphics[width = 1.0\textwidth]{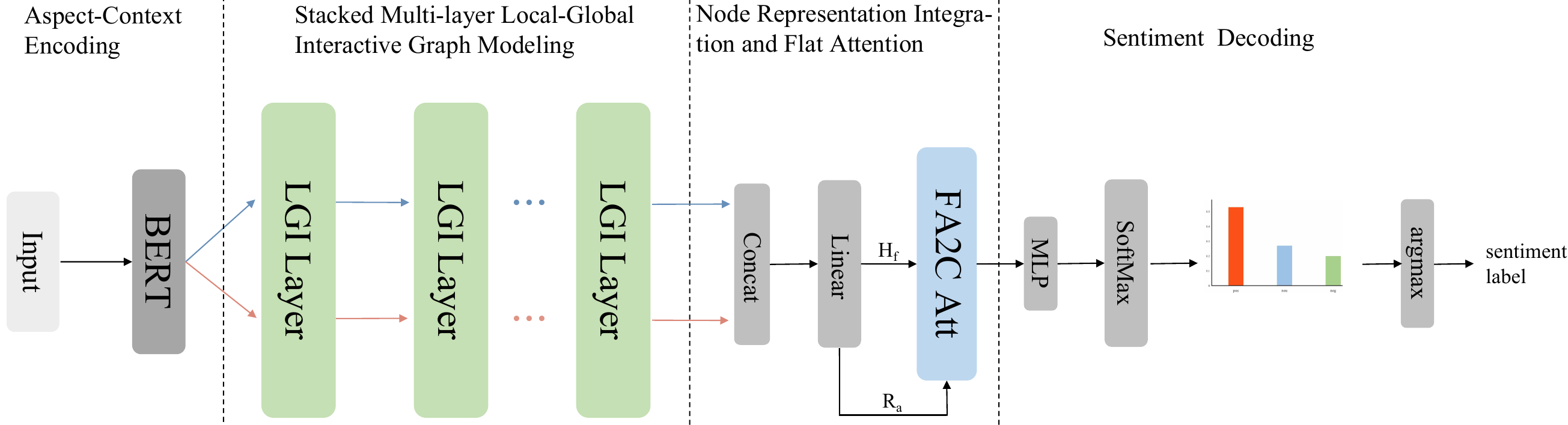}
 \caption{The overall architecture of DigNet.}
 \label{fig: framework}
\end{figure*}

Previous syntax-based ASC models only explore either syntax graph or relation graph to dig the clues for ASC.
As a result, they suffer from the issues caused by the shortages of syntax graph and relation graph, leading to the loss of some crucial clues.
To this end, we propose LGIG to marry syntax graph and relation graph then let them compensate for each other's shortages.
To model and leverage LGIG to address the ASC task, we propose a novel model termed DigNet whose architecture is illustrated in Fig. \ref{fig: framework}.
It is composed of three modules: Aspect-Context Encoding, \textit{Local-Global Interactive Graph Modeling} and Sentiment Decoding.
In this section, we describe the details of different components of DigNet.

\subsection{Aspect-Context Encoding}
We adopt the BERT encoder to generate the hidden state for each context word.
And the input of BERT is the aspect-context pair:
$\langle$[CLS]; context; [SEP]; aspect; [SEP]$\rangle$,
where $\langle;\rangle$ denotes sequence concatenation.
In this way, the BERT encoder is aware of the aspect and then generates aspect-aware context hidden states \cite{AA-LSTM}, which contain not only the intra-sentence dependencies among context words but also the dependencies between the aspect and context.
The aspect hidden states are subsequences of the generated context hidden states.
In both syntax graph and relation graph, the aspect is a single node, so we collapse the aspect hidden states as a single hidden state.
We choose the hidden state of [CLS] $h_{cls}$ as the initial aspect representation then take it to replace the original aspect hidden states in generated context hidden states.
Finally, we obtain a series of hidden state: $\mathbf{H_0} = {[h_c^1, h_c^2, ..., h_{cls}, ..., h_c^{N}]}$, where $N=N_c-N_a+1$, $N_c$ is the number of context words and $N_a$ is the number of the aspect words.


\subsection{Local-Global Interactive Graph Modeling}
\begin{figure}[t]
 \centering
 \includegraphics[width = 0.7\textwidth]{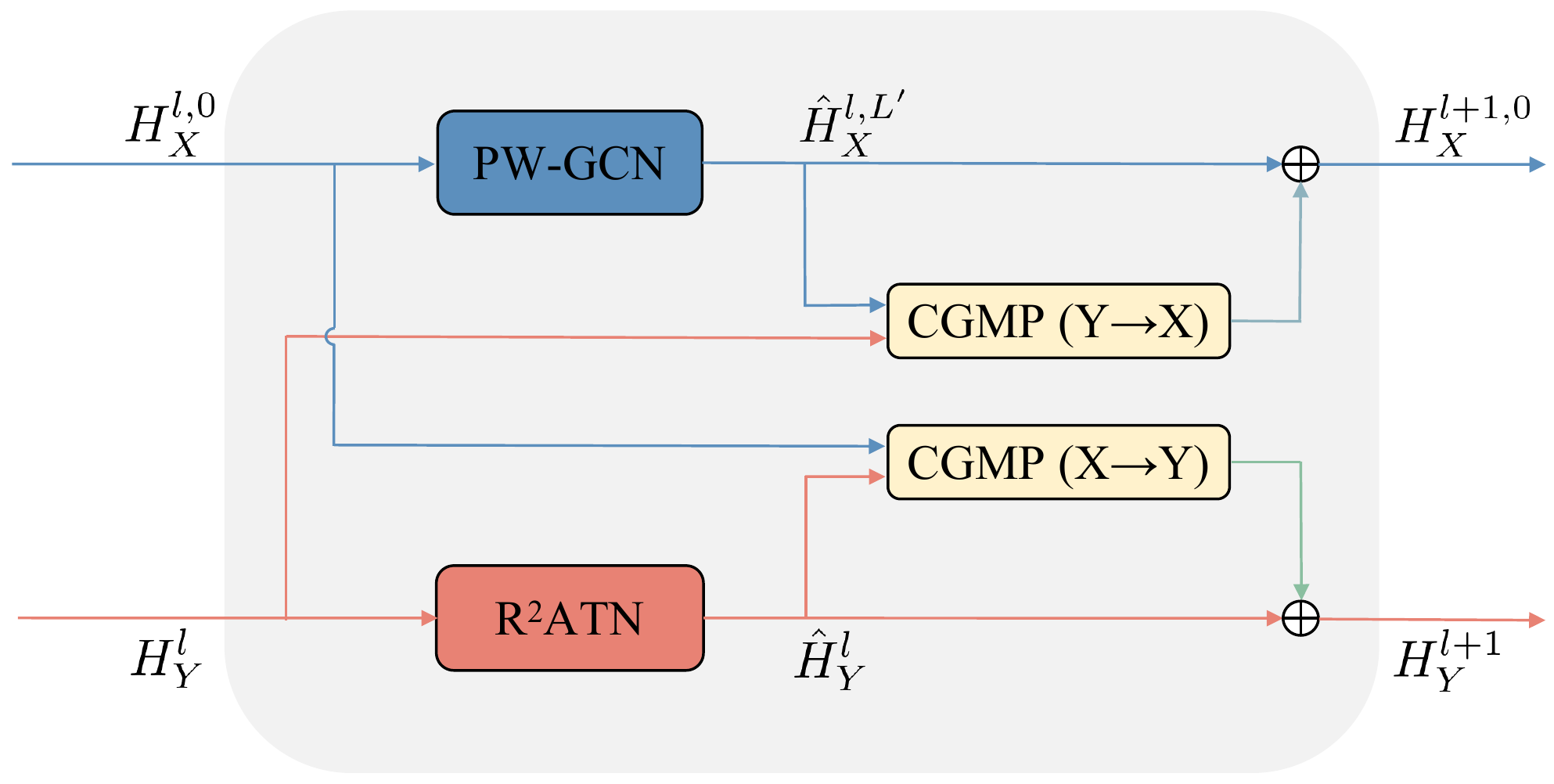}
 \caption{Illustration of local-global interactive layer.}
 \label{fig: LGI}
\end{figure}
This module is composed of multiple stacked local-global (LGI) interactive layers.
The architecture of a single LGI layer is shown in Fig. \ref{fig: dvig}.
LGI layer allows syntax graph and relation graph to share and exchange beneficial information with each other, comprehensively digging sufficient clues for ASC.
It includes two processes: intra-graph message passing (IGMP) and cross-graph message passing (CGMP).
LGI layer has two streams of input: $H_X^{l,0}$ and $H_Y^l$, which denote the hidden states of syntax graph's nodes and relation graph's nodes, respectively.
The input of IGMP is the current graph's hidden states series.
As for CGMP, there are two inputs: 1) another graph's hidden states series from the previous LGI layer; 2) this graph's hidden states series that have been processed by IGMP in the current LGI layer.
At $0-$th LGI layer, the two inputs are the same: $H_0$.
Formally, $l$-th LGI layer can be formulated as:
\begin{align}
h_{x,i}^{l+1,0} = & \operatorname{IGMP}_x^l + \operatorname{CGMP}^{y}_{x}(l)  \label{eq: gcn}\\
h_{y,i}^{l+1} = &\operatorname{IGMP}_y^l + \operatorname{CGMP}^{x}_{y}(l) \label{eq: r2atn}
\end{align}
where $\operatorname{CGMP}^{y}_{x}(l)$ denotes the CGMP (Y$\rightarrow$X) in Fig. \ref{fig: LGI}, indicating the message is transferred from relation graph to syntax graph, and $\operatorname{CGMP}^{x}_{y}(l)$ denotes the CGMP (X$\rightarrow$Y) in Fig. \ref{fig: LGI}, indicating the message is transferred from syntax graph to relation graph.

Concretely, $\operatorname{IGMP}_x^l$ is achieved by a position-weighted GCN (PW-GCN), while $\operatorname{IGMP}_y^l$ is achieved by our proposed reversible relational attention network (R$^2$ATN).
As for CGMP, we design three variants: (1) cross-graph gate (CgGate); (2) cross-graph multi-layer perception (CgMLP); (3) cross-graph multi-head attention (CgMHA).
CgGate and CgMLP operate on O2O-LGIG, and CgMHA operate on O2A-LGIG.
Next, we introduce the details of IGMP and CGMP.

\subsubsection{Intra-graph Message Passing} 
IGMP aggregates information among the nodes in a graph regarding its own structure.
On the syntax graph based on the local syntactic connections, PW-GCN captures local syntactic clues.
On the relation graph based on the global relational connections, R$^2$ATN captures global relational clues.

\noindent \textbf{Position-weighted GCN.} \
Vanilla GCN does not consider the order of nodes, losing the positional information, which is vital for modeling the syntax graph.
We argue that each node's position weight ($w_p$) regarding its offset relative to $v_a^x$ should be introduced into GCN, and there are three reasons.
First, $w_p$ can indicate the position of $v_a^x$ then assist in aggregating the aspect-related semantics.
Second, $w_p$ can highlight the aspect-related nodes which are usually close to $v_a^x$.
Third, the positional information can facilitate GCN's message passing to discover important clues for ASC.
For instance, in Fig. \ref{fig: dual-view} (1), if GCN is aware that `not' is closer to `dosa' than `cheap', it can capture the clue that `dosa' is `not cheap'.

Therefore, in this work we propose a position-weighted GCN (PW-GCN) to introduce node's positional information into GCN and it can be formulated as:
\begin{align}
\hat{h}_{x,i}^{l,l'}=& \sum_{j\in \mathcal{N}_{i}^x} {W_x} (w_p^j\ h_{x,j}^{l,l'-1})  /(d_i + 1) + b_x\\
w_p^i =& 1 - \vert i-\tau\vert/(N+1)
\end{align}
where $\mathcal{N}_{i}^x$ denotes the first-order neighborhoods of $i-$th node on syntax graph,  $d_i$ denotes the degree of $i-$th node, $\tau$ indicate the index of $v_a^x$, $l$ denotes $l$-th LGI layer, and
$l'$ denotes $l'$-th GCN layer.
The output of PW-GCN is $\hat{H}_X^{l,L'}$, where $L'$ is the GCN layer number.

\noindent \textbf{Reversible Relational Attention Network.} \
Operating on relation graph, RGAT \cite{RGAT} aggregates the information from each non-aspect node to $v_a^y$, whose representation is fed to classifier.
However, it cannot be applied to our stacked multi-layer local-global interactive graph modeling since it cannot update non-aspect nodes.
To this end, we propose a reversible relational attention network (R$^2$ATN), which firstly aggregates the message from non-aspect nodes to $v_a^y$, then passes the message of $v_a^y$ to each non-aspect node regarding the reversed relation. 
R$^2$ATN can be formulated as:
\begin{align}
\widetilde{h^{l}_{y,a}}&=\sum_{m=1}^{N^y_h} \Bigg(\sum_{i=1}^{N} \alpha_{i}^m W_{m}^{1} h^{l-1}_{y,i}\Bigg)/N^y_h \\
\alpha_{i}^m& = \operatorname{SoftMax}(\operatorname{FFN} \left(\phi^{emb}(r^{i})\right)\\
\hat{h}^{l}_{y,i}&= h^{l}_{y,i} +\sum_{m=1}^{N^y_h} (\phi^{emb}_m(r^{i}_{rv}) \widetilde{h^{l}_{y,a}} )/N^y_h 
\end{align}
where $N^y_h$ is the number of heads;
 $\operatorname{FFN}$ is the feed-forward network: $\operatorname{FFN}(x) = max(0,xW_1+b_1)W_2+b2$;
 $r_i$ and $r^{i}_{rv}$ denote the relation and reversed relation between $A$ and node $i$;
  $\phi^{emb}(r^{i})$ is the embedding of $r^i$;
  $\phi^{emb}_m(r^{i}_{rv})$ is the embedding of $r^i_{rv}$ corresponding to the $m-$th head and in this work we assign $\phi^{emb}_m(r_{rv}^{i})=\alpha_{i}^m$.

Here $v_a^y$ serves as a message transfer station, which allows each non-aspect word to communicate with others indirectly.
We believe this can not only facilitate modeling the interactions between the syntax graph and relation graph but also help alleviate the isolating issue of the relation graph, enhancing the ability to capture local clues.

\subsubsection{Cross-graph Message Passing}\
CGMP allows the direct communication between syntax graph's nodes and relation graph's nodes.
In doing so, the local syntactic information can be transferred from syntax graph's nodes to relation graph's nodes.
And the global relational information can be transferred from relation graph's nodes to syntax graph's nodes.
Hence, syntax graph and relation graph compensate for each other, and both the local syntactic information and global relational information can flow in each graph.
These comprehensive and sufficient syntactic clues help infer the aspect's sentiment.
Next, we introduce the three variants for cross-graph interactions.

\noindent \textbf{CgGate.}\
 Inspired by the gating mechanism in LSTM \cite{LSTM} and GRU \cite{gru}, we design a CgGate to adaptively determine what and how much information from the counterpart node on another graph should be passed to current node. CgGate can be formulated as:
\begin{align}
\operatorname{CgGate}^y_x(l) &= h_{y,i}^{l} \odot (W_{gate}^{y\rightarrow x} [h_{y,i}^{l}, \hat{h}_{x,i}^{l,L'}])\\
\operatorname{CgGate}^x_y(l) &= h_{x,i}^{l,L'} \odot (W_{gate}^{x\rightarrow y} [h_{x,i}^{l,L'}, \hat{h}_{y,i}^{l}])
\end{align}
where $\odot$ denotes the element-wise multiplication and $[,]$ denotes vector concatenation.

\noindent \textbf{CgMLP.}\
 MLP can automatically learn the integrated representation. Here we employ two MLPs to respectively learn the integrated node representations for syntax graph and relation graph:
\begin{align}
\operatorname{CgMLP}^y_x(l) =& \operatorname{MLP}^y_x([h_{y,i}^{l}, \hat{h}_{x,i}^{l,L'}]) \\
\operatorname{CgMLP}^x_y(l) =& \operatorname{MLP}^x_y([h_{x,i}^{l,L'}, \hat{h}_{y,i}^{l}])
\end{align}
\noindent \textbf{CgMHA.}\
 On O2A-LGIG, each node is connected to all nodes in another graph.
Based on the multi-head attention (MHA) \cite{transformer}, we propose a CgMHA which takes the current node as query then selectively attends to the related nodes on another graph.
Our CgMHA can be formulated as:
\begin{align}
\operatorname{CgMHA}^y_x(l) =& \operatorname{MHA}^y_x(\hat{H}^{l,L'}_X, H^{l}_Y,H^{l}_Y)\\
\operatorname{CgMHA}^x_y(l) =& \operatorname{MHA}^x_y(\hat{H}^{l}_Y, H^{l,0}_X,H^{l,0}_X)\\
\operatorname{MHA}(Q,K,V) = & \|_{h=1}^{N^M_h}\operatorname{SoftMax}\left(\frac{q k^{\mathsf{T}}}{\sqrt{d_{s}}}\right) v\\
q,k,v=&LT_q(Q), LT_k(K), LT_v(V)
\end{align}
where $LT$ denotes the liner transformation to form $N^M_h$ heads, $\|_{h=1}^{N^M_h}$ denotes the $N^M_h$ outputs are concatenated to form the 1-head output, $d_s$ is the embedding dim of a single head.

\subsection{Sentiment Decoder}
\textbf{Node Representation Integration.}\
After multiple LGI layers, the two streams of node representations from both graphs are integrated into a single series of node representations which contains the important clues comprehensively dug from both graphs: 
\begin{align}
H_{f} =& W_{f}([H_X, H_Y])
\end{align}

\noindent \textbf{Flat Aspect-to-Context Attention.}\
Now we have the final flat series of context word representations $H_f$ and final aspect representation $R_a$.
We utilize a flat aspect-to-context attention (FA2C Att), which takes $R_a$ to query $H_f$ for extracting the aspect-related semantics. FA2C Att can be formulated as:
\begin{align}
R=&\sum_{i=1}^{N}\beta_i h_f^i\\
\beta_i=&\operatorname{SoftMax} (\mathcal{F}(h_f^i, R_a)) \\
\mathcal{F}(h_f^i, R_a)= & (W_{a2c}\,{h_f^i} +b_{a2c})\,R_a^\mathsf{T}
\end{align}
where $\mathsf{T}$ denotes transposition and $R$ is the final sentiment representation.

\noindent \textbf{Prediction and Training.}
We pass $R$ through a MLP then a SoftMax classifier to get the predicted sentiment label:
\begin{align}
P =& \operatorname{SoftMax}(\operatorname{MLP}(R))\\
\widehat{y}=& \operatorname{argmax}_{k\in S}(P[k])
\end{align}
For the training of DigNet, we employ the standard cross-entropy loss as objective function:
\begin{equation}
\mathcal{L}(\theta)=-\sum_{i=1}^{D} \sum_{t=1}^{T} \log P_{i, t}\left[y_{i, t}\right]
\end{equation}
where $\theta$ denotes the parameters of DigNet, $D$ denotes the number of samples, and $T$ denotes the number of aspects in the $i$-th sample.

%% file: experiment.tex
\section{Experiment}\label{sec:experiment}
\subsection{Experimental Setup}
\subsubsection{Dataset}
We conduct experiments on three public benchmark datasets to obtain reliable and authoritative results.
Res14 and Lap14 are from \cite{Semeval2014}, and Res15 is from \cite{semeval2015}.
We pre-process the datasets following the same way as previous works \cite{asgcn,RGAT,tgcn}.
The statistics of all datasets are shown in Table \ref{table: dataset}.

\begin{table}[ht]
\centering
\caption{Dataset statistics of the three datasets.}
\fontsize{10}{12}\selectfont
\setlength{\tabcolsep}{2mm}{
\begin{tabular}{ccccccc}
\toprule
\multirow{2}{*}{Dataset} & \multicolumn{2}{c}{Positive} & \multicolumn{2}{c}{Neutral} & \multicolumn{2}{c}{Negative} \\\specialrule{0em}{0pt}{1pt} \cline{2-3} \cline{4-5} \cline{6-7}\specialrule{0em}{1pt}{1.5pt}
                         & Train         & Test         & Train         & Test        & Train         & Test         \\ \specialrule{0em}{0pt}{1pt}\hline \specialrule{0em}{1.5pt}{1.5pt}
Lap14                    & 994           & 341          &  464          &    169      &  870          &    128       \\
Res14                    & 2164          & 728          & 637           & 196         & 807           & 196          \\
Res15                    & 912           & 326          & 36            & 34          & 256           & 182         \\\bottomrule
\end{tabular}}
\label{table: dataset}
\end{table}

\subsubsection{Implementation Details}
We adopt the BERT-base uncased version as the BERT encoder, which is fine-tuned in the experiments\footnote{ Although larger pre-trained language models can lead to better performances \cite{tgcn}, this is not our focus.}. We train our DigNets using AdamW optimizer \cite{weightdecay}.
Accuracy and Macro-F1 are used as evaluation metrics.
As there is no official validation set for the datasets, following previous works \cite{asgcn,bigcn}, we conventionally report the average results over three runs with random initialization.
For fair comparisons with previous works, we eliminate the impact of dependency parsers:
for syntax graph, following \cite{asgcn,DGEDT,bigcn}, we use spaCy toolkit\footnote{https://spacy.io/.} to obtain the original syntactic structure;
for relation graph, following \cite{RGAT}, we use the Biaffine Parser \cite{biaffine} to obtain the dependency labels.

In our experiments, the dimensions of hidden units and the relation embedding in R$^2$ATN are 768 and 300, respectively.
The dropout rate for the BERT encoder is 0.1, while it is set to 0.3 for other modules.
The batch size is 32 and the epoch number is 30.
The learning rates are $1e-5, 5e-5, 3e-5$ for Lap14, Res14 and Res15 datasets respectively.
The weight decay is $0.05$ for Res14 and Res15 datasets while $0.001$ for Lap14 datasets.
The numbers of PW-GCN and LGI layers are investigated in Sec. \ref{sec:gcn} and Sec. \ref{sec: LGI} respectively.
All of the computations are conducted on an NVIDIA RTX 6000 GPU.
And our source code will be released later.

\subsubsection{Baselines for Comparison}
We choose some models as baselines, which can be divided into four categories regarding whether adopting pre-trained language models (e.g. BERT) and whether utilizing syntax/relation graphs:\\
(1) BERT $\times$ Syntax $\times$:
\begin{itemize}
\item IAN \cite{IAN}: This model separately encodes the aspect and context, then models their interactions through an interactive attention mechanism.
\item RAM \cite{Tencent}: This model uses a GRU attention mechanism to recurrently extracts the aspect-related semantics.
\end{itemize}
(b) BERT $\times$ Syntax $\checkmark$:
\begin{itemize}
\item ASGCN \cite{asgcn}: This model utilizes a GCN to encode the syntax graph for capturing local syntactic information.
\item LSTM+synATT \cite{synatt}: This model leverages syntactic information to improve the conventional attention-based LSTM.
\item BiGCN \cite{bigcn}: This model employs the GCN architecture to convolutes over hierarchical syntactic and lexical graphs.
\end{itemize}
(c) BERT $\checkmark$ Syntax $\times$:
\begin{itemize}
\item BERT-SPC \cite{bert}: This model takes the concatenated aspect-context pair as the input of BERT and the hidden state of [CLS] token is used for classification.
\item AEN-BERT \cite{aen-bert}: This model employs several attention layers to capture aspect-related semantics.
\end{itemize}
(d) BERT $\checkmark$ Syntax $\checkmark$:
\begin{itemize}
\item KGCapsAN-BERT \cite{kgcap}: 
This model utilizes multi-prior knowledge to guide the capsule attention process. Besides, a GCN-based syntactic layer is designed to integrate the syntactic knowledge.
\item ASGCN+BERT \cite{asgcn}: This model replaces the original BiLSTM in ASGCN model with a BERT encoder.
\item R-GAT+BERT \cite{RGAT}: This model includes a relational graph attention network which aggregates the global relational information from all context words into the aspect node representation.
\item DGEDT-BERT \cite{DGEDT}: This model employs a dual-transformer network
to model the interactions between the flat textual knowledge and dependency graph empowered knowledge.
\item KVMN+BERT \cite{kvmn-eacl}: This model uses a key-value memory network
to leverage not only word-word relations but also their dependency types.
\item BERT+T-GCN \cite{tgcn}: This model leverages the dependency types in T-GCN and uses an attentive layer ensemble to learn the comprehensive representation from different T-GCN layers.
\end{itemize}
Note that in all baselines the BERT encoder version is BERT-base uncased version, which is the same as ours.

\subsection{Main Results}

\begin{table*}[ht]
\fontsize{10}{12}\selectfont
\centering
\caption{Performances comparison. Best results are in \textbf{bold} and previous SOTA results are \underline{underlined}. Our DigNets outperform previous SOTA models and corresponding baselines on all datasets, being statistically significant under t-test: $\dag$ denotes $p < 0.01$ and $\ddag$ denotes $p<0.05$.}
\setlength{\tabcolsep}{2.5mm}{
\begin{tabular}{c|cccccc}\toprule
\multirow{2}{*}{Models} & \multicolumn{2}{c}{Lap14} & \multicolumn{2}{c}{Res14} & \multicolumn{2}{c}{Res15}  \\\cline{2-7} 
                                                                       &Acc       &F1        &Acc        &F1        &Acc        &F1       \\\midrule
IAN &72.05     & 67.38    &79.26    &70.09  & 78.54    & 52.65\\
RAM  &74.49& 71.35 & 80.23 &70.80&79.30&60.49\\ \hline
 LSTM+synATT  &72.57 &69.13  & 80.45 & 71.26 & 80.28&65.46 \\ 
 ASGCN  & 75.55    &71.05     &80.77    & 72.02  & 79.89   & 61.89\\
 BiGCN  & 74.59    &71.84    &81.97     &73.48    & 81.16    & 64.79  \\ \hline
 BERT-SPC & 78.47    &73.67  &84.94   & 78.00   & 83.40    &  65.00   \\
 AEN-BERT  & 79.93    &76.31  &83.12   & 73.76   & -    &  -   \\ \hline
  KGCapsAN-BERT   & 79.47    &76.61    &85.36      & 79.00   & - & - \\
 ASGCN+BERT     & 78.92    &74.35     &85.87    & 79.32  & 83.85   & 68.73        \\
R-GAT+BERT & 79.31    &75.40  &86.10   & \underline{80.04}   & 83.95    &  69.47   \\
DGEDT-BERT   & 79.8    &75.6    &\underline{86.3}      & 80.0   & 84.0   &71.0    \\
 A-KVMN+BERT   & 79.20    &75.76   &85.89      & 78.29   & 83.89  &67.88    \\
 BERT+T-GCN    & \underline{80.56}    &\underline{76.95}    &85.95      & 79.40   & \underline{84.81} & \underline{71.09} \\
\midrule \midrule
  DigNet-CgGate   &\textbf{82.24}$^\dag$ & \textbf{79.26}$^\dag$  &87.08$^\dag$  & 81.45$^\dag$ &\textbf{86.29}$^\dag$  &75.20$^\dag$     \\
 DigNet-CgMHA   &82.03$^\dag$ & 78.93$^\dag$ &\textbf{87.44}$^\dag$   & \textbf{81.89}$^\dag$ &86.04$^\dag$ &75.58$^\dag$   \\
  DigNet-CgMLP   &81.19$^\dag$    & 78.00$^\dag$    &87.05$^\dag$   & 81.23$^\ddag$  &85.79$^\ddag$ &\textbf{75.59}$^\dag$        \\
 \textbf{Improve} & \textbf{2.09\%}  & \textbf{3\%}  &\textbf{1.32\%}   & \textbf{2.32\%}  &\textbf{1.75\%}  &\textbf{6.33\%} \\
\bottomrule
\end{tabular}}
\label{table: results}
\end{table*}

The performances comparisons of all models are shown in Table \ref{table: results}.
Our DigNets achieve the best performance on all datasets and surpass previous state-of-the-art models with significant improvements on both Accuracy (Acc) and Macro-F1 (F1). 

From the table, we get an overview of how ASC models evolve chronologically. 
The first category models employ the RNN + Attention architecture, and they perform worst since that they do not leverage the syntactic information of the context or BERT encoder.
As the syntactic information is utilized, the second category models can capture some important clues which are lost by attention mechanisms.
Consequently, the second category models obviously outperform the first category models.
Then we can find that employing the BERT encoder can significantly improve the performances. And even the simple BERT-SPC model can overpass all the models in the first and second categories. This is because pre-trained BERT has a strong ability to capture the semantic dependencies between the tokens in a sequence. And we can observe that the BERT encoder and the syntactic information can cooperate with each other, further boosting the performances. 

In the fourth category models, most of them use the syntax graph which is based on local syntactic connections, while R-GAT+BERT proposes and utilizes the relation graph which is based on global relational connections.
Although they have obtained promising results, they 
only utilize either syntax graph or relation graph, whose intrinsic shortages are neglected while limiting the performances, leaving space for improvement.
In this paper, we find that the advantages of syntax graph and relation graph can make up for each other's shortages.
Inspired by this, we first design the local-global interactive graph (LGIG) which marries syntax graph and relation graph by interactive edges, then propose DigNet to leverage LGIG and comprehensively dig sufficient clues from it.
As a result, we obtain consistent improvements over all baselines in terms of both Acc and F1.
Our DigNets surpass previous best scores by 2.09\%, 1.32\%, and 1.75\% in terms of Acc on Lap14, Res14, and Res15 datasets, respectively.
And in terms of F1, our DigNets surpass previous best scores by 3\%, 2.32\%, and 6.33\% on Lap14, Res14, and Res15 datasets, respectively.

\subsection{Ablation Study} \label{sec: ablate}
Based on DigNet-CgMHA, we conduct a set of analysis experiments on Lap14 and Res14 datasets to prove the necessity of each component of DigNet and interpret the efficacy of LGI which is the core of this work.
In this section, we answer the following research questions (\textbf{RQs}) based on the experimental results listed in Table \ref{table: model analysis}.

\textbf{RQ1}: \textit{Are syntax graph and relation graph both indispensable?}

To answer this question, we design two variants: 
w/o syntax denotes that we remove PW-GCN from DigNet and delete syntax graph from LGIG;
w/o relation denotes that we remove R$^2$ATN from DigNet and delete the relation graph from LGIG.
We can observe that only utilizing syntax graph or relation graph leads to a significant performance decrease.
Therefore, both the syntax graph and relation graph are beneficial for ASC and crucial for DigNet.

\begin{table*}[t]
\fontsize{10}{12}\selectfont
\centering
\caption{Experimental results for model analysis.}
\setlength{\tabcolsep}{2mm}{
\begin{tabular}{c|cc|cc}\toprule
\multirow{2}{*}{Model} & \multicolumn{2}{c|}{Lap14} & \multicolumn{2}{c}{Res14}  \\ \cline{2-5}
                                     &Acc       &F1        &Acc        &F1             \\\midrule
 DigNet-CgMHA   &82.03 & 78.93                  & 87.44                       & 81.89    \\\hline   
w/o syntax  & 79.94 ($\downarrow$2.09)  &  76.33 ($\downarrow$2.60)  &  86.64  ($\downarrow$0.80)  & 80.61 ($\downarrow$1.28)   \\
w/o relation  & 79.89 ($\downarrow$2.14) & 76.21 ($\downarrow$2.72)  & 86.55 ($\downarrow$0.89)  & 80.51 ($\downarrow$1.38)\\
w/o LGI     &80.00  ($\downarrow$2.03)  & 76.61  ($\downarrow$2.32)  &86.40 ($\downarrow$1.04) & 80.13  ($\downarrow$1.76) \\
w/o FA2C Att &  81.09 ($\downarrow$0.94) & 77.54 ($\downarrow$1.39) & 86.93 ($\downarrow$0.51) & 81.22 ($\downarrow$0.67)\\ 
syntax-decoder  & 81.50 ($\downarrow$0.53)  &  78.28 ($\downarrow$0.65) &  86.82 ($\downarrow$0.62) & 81.06 ($\downarrow$0.83)\\
relation-decoder  &  81.71 ($\downarrow$0.32) &  78.21 ($\downarrow$0.72) &  87.05 ($\downarrow$0.39) & 81.38 ($\downarrow$0.51)\\
\bottomrule
\end{tabular}}
\label{table: model analysis}
\end{table*}

\textbf{RQ2}: \textit{What is the impact of LGI layers on performances?}

To answer this question, we design the w/o LGI variant by replacing the LGI layers with independent PW-GCN and $R^2$ATN, whose outputs are then fed to Sentiment Decoding module.
In this case, there is no interaction between the nodes on the syntax graph and the ones on the relation graph.
From the inferior performance of w/o LGI, we can conclude that LGI is effective in significantly improving ASC performance via allowing the nodes on different graphs to exchange beneficial information with each other.

\textbf{RQ3}: \textit{Is the flat aspect-to-context attention (FA2C Att) necessary?}

To answer this question, we remove FA2C Att and use the final aspect representation $R_a$ for classification.
We can find that the variant w/o FA2C Att performs worse than DigNet-CgMHA.
We speculate that there are two reasons.
First, the clues conveyed by $R_a$ are not enough and there are some important sentiment clues contained in the hidden states of non-aspect words.
And FA2C Att can extract these important semantics and aggregate them into the final sentiment representation $R$.
Second, the original syntax graphs obtained by dependency parsers may be imperfect \cite{DGEDT}, so some wrong connections or dependency labels are generated.
In this case, node representations output by LGI layers may contain some noise information, which can be alleviated by FA2C Att.


\textbf{RQ4}: \textit{Why LGI layers can improve ASC performances?}

To interpret how LGI layers improve performances, we design two variants: syntax-decoder and relation-decoder.
Syntax-decoder denotes that after multiple LGI layers the nodes representations of syntax graph ($H_X$) are taken as the final hidden states $H_f$.
And relation-decoder denotes that after multiple LGI layers, the nodes representations of relation graph ($H_Y$) are taken as $H_f$.
The similarity of syntax-decoder and w/o relation is only using syntax graph's nodes representations for prediction, while their difference is that syntax-decoder leverages relation graph, but w/o relation does not.
We can find that syntax-decoder significantly outperforms w/o relation.
This can prove that the nodes representations in syntax-decode are much better than the ones in w/o relation, containing better and more sufficient clues for ASC.
Hence we can conclude that the LGI layers in syntax-decoder can improve the nodes representations of syntax graph via leveraging relation graph.
And similar conclusion can be made by comparing relation-decoder and w/o syntax.
Therefore, the reason why LGI layers can improve ASC performances is that they can improve the nodes representations of both graphs by allowing syntax graph and relation graph to compensate for each other's shortages using their own advantages. 

\textbf{RQ5}: \textit{Is the node representation integration necessary?}

As syntax-decoder and relation-decoder respectively take $H_X$ and $H_Y$ as $H_f$, there is no node representation integration in these two variants.
We can find that their performances are both inferior to DigNet-CgMHA.
We speculate the reason is that after multi-layer LGI, the nodes representations of syntax graph contain specific clues based on the local syntactic connections, and the nodes representations of relation graph contain specific clues based on the global relational connections.
Both of the two streams of nodes representations should be considered together since that the clues conveyed in either one stream of nodes representations are not enough.
Consequently, the node representation integration is necessary to combine the representations of both graphs to obtain sufficient clues for ASC.

\subsection{Effect of LGI Layers Number}\label{sec: LGI}
\begin{figure*}[t]
 \centering
 \includegraphics[width = \textwidth]{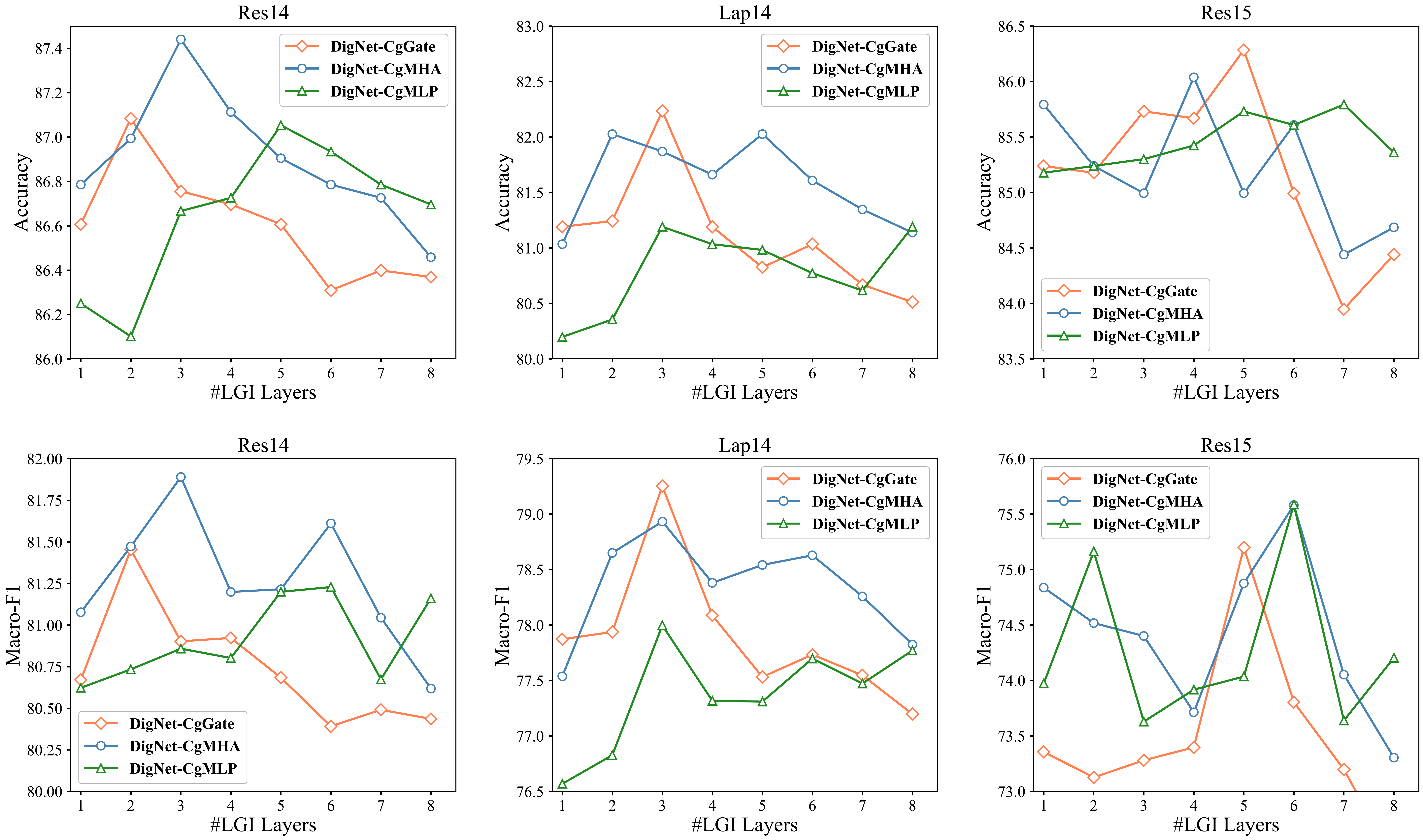}
 \caption{Test results of DigNet-CgGate, DigNet-CgMHA and DigNet-CgMLP on the three datasets by different numbers of LGI layers. All scores are the average of three runs with random initialization.}
 \label{fig: dvi-layer}
\end{figure*}
We plot the performance trends of DigNet-CgGate, DigNet-CgMHA, and DigNet-CgMLP on the three datasets, as presented in Fig. \ref{fig: dvi-layer}.
We can find that as the number of LGI layers increases, the performances show a trend of increases at first and then decreases.
And generally, the best performances are obtained when the number is 2 or 3 for Lap14 and Res14 dataset, while about 5 for Res15 dataset.
We suspect the reason why Res15 dataset demands a larger number of LGI layers is that Res15 consists of much fewer samples than Res14 and Lap14, then DigNet tries to make the most of each sample by digging the clues in more LGI layers.
Overall, the phenomenon that appropriately increasing LGI layers number can gradually improve the results is consistent with our expectations.
In the stacking of multiple LGI layers, syntax graph nodes and relation graph nodes can gradually transfer beneficial information to each other, then both local syntactic clues and global relation clues can be discovered and leveraged for ASC.
This can also prove the necessity of the multi-layer stacking manner of LGI.
However, we can find that too large LGI layers number leads to performance decrease, which is also consistent with our expectation. One possible reason is too many LGI layers lead to too deep information fusion of the nodes on syntax graph and the ones on relation graph, resulting in the loss of respective specific key clues which are originally captured based on their individual characteristics.
Another possible reason is that too many LGI layers lead to overfitting on training sets, resulting in worse results on test sets.

\subsection{Effect of PW-GCN Layers Number}\label{sec:gcn}
\begin{table}[ht]
\fontsize{10}{12}\selectfont
\centering
\caption{The PW-GCN layers number settings of best results. }
\setlength{\tabcolsep}{2mm}{
\begin{tabular}{c|ccc}\toprule
Model & Lap14 & Res14 & Res15  \\ \midrule
DigNet-CgGate &2 & 2 & 2 \\ \hline
DigNet-CgMHA & 3 & 2 & 2 \\ \hline
DigNet-CgMLP &2 & 2 & 3 \\ 
\bottomrule
\end{tabular}}
\label{table: gcn-layer}
\end{table}

\begin{figure}[h]
 \centering
 \includegraphics[width = 0.8\textwidth]{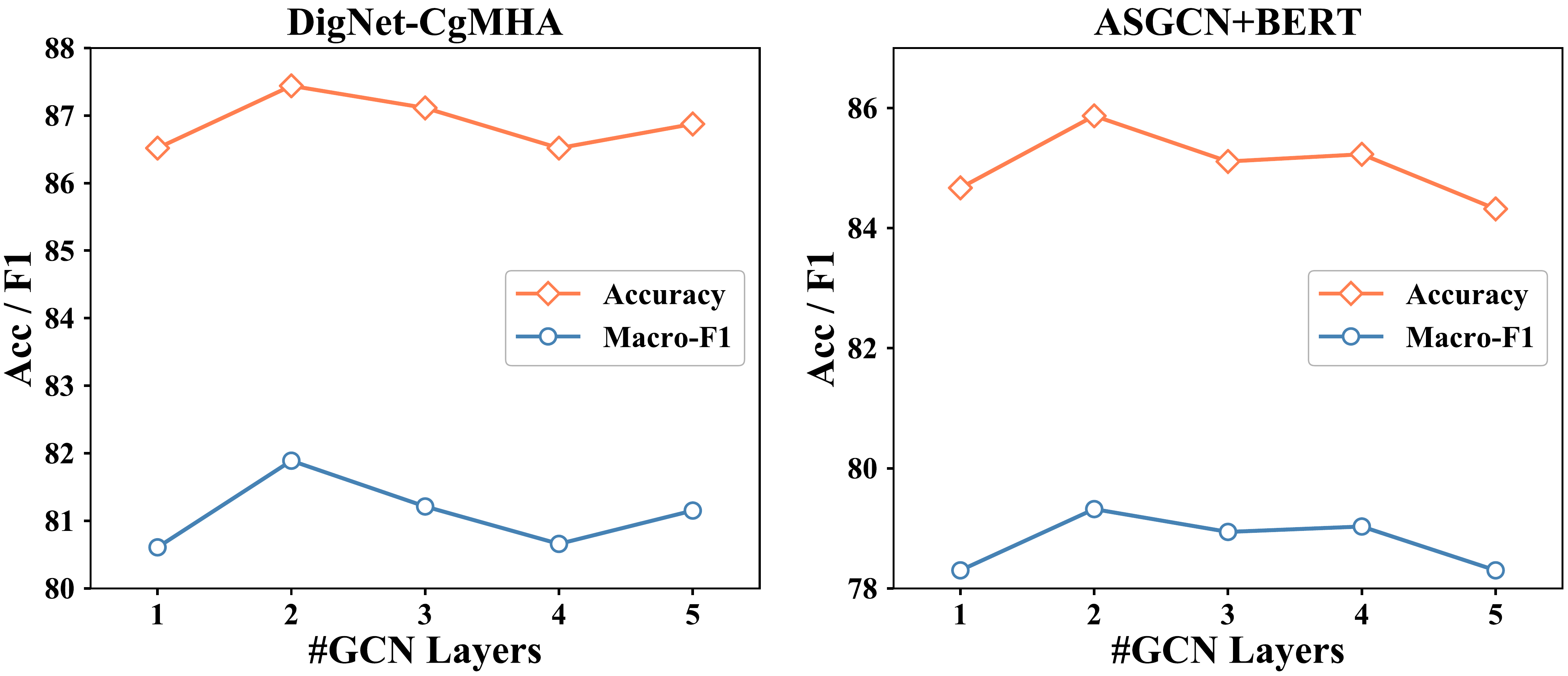}
 \caption{Test results of DigNet-CgMHA on the Lap14 and Res14 dataset by different numbers of PW-GCN layers.}
 \label{fig: gcn-layer}
\end{figure}

We list the optimal numbers of PW-GCN layers for DigNets on the three datasets in Table \ref{table: gcn-layer}.
Generally, our DigNets perform best when PW-GCN has two layers, which is consistent with the optimal GCN layer number in previous works \cite{asgcn,CDT,bigcn}.
And we examine the effect of PW-GCN layer number (from 1 to 5) on DigNet-CgMHA and also take ASGCN-BERT for comparison.
The results are shown in Fig. \ref{fig: gcn-layer}.
We can find that only adopting one layer of GCN, which means that only 1-hop message passing can be achieved, cannot sufficiently capture local clues. 
However, if continue increase the layer number larger than 2, the performances do not get improved.
This is because a large layer number introduces more parameters and makes it more difficult to train the model.

Then we can come to the conclusion that the reason why DigNets significantly outperform previous works is that it allows syntax graph and relation graph compensate for each other and improve nodes representations by cross-graph message passing, rather than increasing the distance of message passing of GCN.

%% file: conclusion.tex
\section{Conclusion}\label{sec: conclusion}
In this paper, we point out the respective shortages of syntax graph and relation graph.
And we discover that their advantages can compensate for each other's shortages.
To this end, we design a novel local-global interactive graph (LGIG) to marry syntax graph and relation graph which are based on local syntactic connections and global relational connections, respectively. 
To encode and leverage LGIG, we propose the DigNet model which simultaneously captures local syntactic information and global relational information, then allows them compensate for each other.
Hence DigNet can comprehensively dig sufficient clues from LGIG and
experimental results show that our DigNet achieves new state-of-the-art results.
To the best of our knowledge, our DigNet represents the first attempt to combine the local syntactic information and global relational information, providing a strong baseline for ASC community.

%% file: dignet_paper.bbl
\begin{thebibliography}{}

\bibitem[\protect\BCAY{Bai, Liu,\ \BBA\ Zhang}{Bai et~al.}{2020}]{yuezhangRGAT}
Bai, X., Liu, P., \BBA\ Zhang, Y. \BBOP2020\BBCP.
\newblock \BBOQ Investigating typed syntactic dependencies for targeted
  sentiment classification using graph attention neural network\BBCQ\
\newblock {\Bem IEEE/ACM Transactions on Audio, Speech, and Language
  Processing}.

\bibitem[\protect\BCAY{Chen, Sun, Bing,\ \BBA\ Yang}{Chen
  et~al.}{2017}]{Tencent}
Chen, P., Sun, Z., Bing, L., \BBA\ Yang, W. \BBOP2017\BBCP.
\newblock \BBOQ Recurrent attention network on memory for aspect sentiment
  analysis\BBCQ\
\newblock In {\Bem Proceedings of the 2017 Conference on Empirical Methods in
  Natural Language Processing}, \BPGS\ 452--461, Copenhagen, Denmark.
  Association for Computational Linguistics.

\bibitem[\protect\BCAY{Cho, van Merri{\"e}nboer, Gulcehre, Bahdanau, Bougares,
  Schwenk,\ \BBA\ Bengio}{Cho et~al.}{2014}]{gru}
Cho, K., van Merri{\"e}nboer, B., Gulcehre, C., Bahdanau, D., Bougares, F.,
  Schwenk, H., \BBA\ Bengio, Y. \BBOP2014\BBCP.
\newblock \BBOQ Learning phrase representations using {RNN} encoder{--}decoder
  for statistical machine translation\BBCQ\
\newblock In {\Bem Proceedings of the 2014 Conference on Empirical Methods in
  Natural Language Processing ({EMNLP})}, \BPGS\ 1724--1734, Doha, Qatar.
  Association for Computational Linguistics.

\bibitem[\protect\BCAY{Devlin, Chang, Lee,\ \BBA\ Toutanova}{Devlin
  et~al.}{2019}]{bert}
Devlin, J., Chang, M.-W., Lee, K., \BBA\ Toutanova, K. \BBOP2019\BBCP.
\newblock \BBOQ {BERT}: Pre-training of deep bidirectional transformers for
  language understanding\BBCQ\
\newblock In {\Bem Proceedings of the 2019 Conference of the North {A}merican
  Chapter of the Association for Computational Linguistics: Human Language
  Technologies, Volume 1 (Long and Short Papers)}, \BPGS\ 4171--4186,
  Minneapolis, Minnesota. Association for Computational Linguistics.

\bibitem[\protect\BCAY{Dozat\ \BBA\ Manning}{Dozat\ \BBA\
  Manning}{2017}]{biaffine}
Dozat, T.\BBACOMMA\  \BBA\ Manning, C.~D. \BBOP2017\BBCP.
\newblock \BBOQ Deep biaffine attention for neural dependency parsing\BBCQ\
\newblock In {\Bem 5th International Conference on Learning Representations,
  {ICLR} 2017, Toulon, France, April 24-26, 2017, Conference Track
  Proceedings}. OpenReview.net.

\bibitem[\protect\BCAY{He, Lee, Ng,\ \BBA\ Dahlmeier}{He et~al.}{2018}]{synatt}
He, R., Lee, W.~S., Ng, H.~T., \BBA\ Dahlmeier, D. \BBOP2018\BBCP.
\newblock \BBOQ Effective attention modeling for aspect-level sentiment
  classification\BBCQ\
\newblock In {\Bem Proceedings of the 27th International Conference on
  Computational Linguistics}, \BPGS\ 1121--1131, Santa Fe, New Mexico, USA.
  Association for Computational Linguistics.

\bibitem[\protect\BCAY{Hochreiter\ \BBA\ Schmidhuber}{Hochreiter\ \BBA\
  Schmidhuber}{1997}]{LSTM}
Hochreiter, S.\BBACOMMA\  \BBA\ Schmidhuber, J. \BBOP1997\BBCP.
\newblock \BBOQ Long short-term memory\BBCQ\
\newblock {\Bem Neural computation}, {\Bem 9\/}(8), 1735--1780.

\bibitem[\protect\BCAY{Huang\ \BBA\ Carley}{Huang\ \BBA\ Carley}{2018}]{pcnn}
Huang, B.\BBACOMMA\  \BBA\ Carley, K. \BBOP2018\BBCP.
\newblock \BBOQ Parameterized convolutional neural networks for aspect level
  sentiment classification\BBCQ\
\newblock In {\Bem Proceedings of the 2018 Conference on Empirical Methods in
  Natural Language Processing}, \BPGS\ 1091--1096, Brussels, Belgium.
  Association for Computational Linguistics.

\bibitem[\protect\BCAY{Huang\ \BBA\ Carley}{Huang\ \BBA\
  Carley}{2019}]{graphatt}
Huang, B.\BBACOMMA\  \BBA\ Carley, K. \BBOP2019\BBCP.
\newblock \BBOQ Syntax-aware aspect level sentiment classification with graph
  attention networks\BBCQ\
\newblock In {\Bem Proceedings of the 2019 Conference on Empirical Methods in
  Natural Language Processing and the 9th International Joint Conference on
  Natural Language Processing (EMNLP-IJCNLP)}, \BPGS\ 5469--5477, Hong Kong,
  China. Association for Computational Linguistics.

\bibitem[\protect\BCAY{Huang, Sun, Li, Zhang,\ \BBA\ Wang}{Huang
  et~al.}{2020}]{sagat}
Huang, L., Sun, X., Li, S., Zhang, L., \BBA\ Wang, H. \BBOP2020\BBCP.
\newblock \BBOQ Syntax-aware graph attention network for aspect-level sentiment
  classification\BBCQ\
\newblock In {\Bem Proceedings of the 28th International Conference on
  Computational Linguistics}, \BPGS\ 799--810, Barcelona, Spain (Online).
  International Committee on Computational Linguistics.

\bibitem[\protect\BCAY{Kipf\ \BBA\ Welling}{Kipf\ \BBA\ Welling}{2017}]{gcn}
Kipf, T.~N.\BBACOMMA\  \BBA\ Welling, M. \BBOP2017\BBCP.
\newblock \BBOQ Semi-supervised classification with graph convolutional
  networks\BBCQ\
\newblock In {\Bem 5th International Conference on Learning Representations,
  {ICLR} 2017, Toulon, France, April 24-26, 2017, Conference Track
  Proceedings}. OpenReview.net.

\bibitem[\protect\BCAY{Loshchilov\ \BBA\ Hutter}{Loshchilov\ \BBA\
  Hutter}{2019}]{weightdecay}
Loshchilov, I.\BBACOMMA\  \BBA\ Hutter, F. \BBOP2019\BBCP.
\newblock \BBOQ Decoupled weight decay regularization\BBCQ\
\newblock In {\Bem 7th International Conference on Learning Representations,
  {ICLR} 2019, New Orleans, LA, USA, May 6-9, 2019}. OpenReview.net.

\bibitem[\protect\BCAY{Ma, Li, Zhang,\ \BBA\ Wang}{Ma et~al.}{2017}]{IAN}
Ma, D., Li, S., Zhang, X., \BBA\ Wang, H. \BBOP2017\BBCP.
\newblock \BBOQ Interactive attention networks for aspect-level sentiment
  classification\BBCQ\
\newblock In Sierra, C.\BED, {\Bem Proceedings of the Twenty-Sixth
  International Joint Conference on Artificial Intelligence, {IJCAI} 2017,
  Melbourne, Australia, August 19-25, 2017}, \BPGS\ 4068--4074. ijcai.org.

\bibitem[\protect\BCAY{Pontiki, Galanis, Papageorgiou, Manandhar,\ \BBA\
  Androutsopoulos}{Pontiki et~al.}{2015}]{semeval2015}
Pontiki, M., Galanis, D., Papageorgiou, H., Manandhar, S., \BBA\
  Androutsopoulos, I. \BBOP2015\BBCP.
\newblock \BBOQ {S}em{E}val-2015 task 12: Aspect based sentiment analysis\BBCQ\
\newblock In {\Bem Proceedings of the 9th International Workshop on Semantic
  Evaluation ({S}em{E}val 2015)}, \BPGS\ 486--495, Denver, Colorado.
  Association for Computational Linguistics.

\bibitem[\protect\BCAY{Pontiki, Galanis, Pavlopoulos, Papageorgiou,
  Androutsopoulos,\ \BBA\ Manandhar}{Pontiki et~al.}{2014}]{Semeval2014}
Pontiki, M., Galanis, D., Pavlopoulos, J., Papageorgiou, H., Androutsopoulos,
  I., \BBA\ Manandhar, S. \BBOP2014\BBCP.
\newblock \BBOQ {S}em{E}val-2014 task 4: Aspect based sentiment analysis\BBCQ\
\newblock In {\Bem Proceedings of the 8th International Workshop on Semantic
  Evaluation ({S}em{E}val 2014)}, \BPGS\ 27--35, Dublin, Ireland. Association
  for Computational Linguistics.

\bibitem[\protect\BCAY{Schouten\ \BBA\ Frasincar}{Schouten\ \BBA\
  Frasincar}{2015}]{tkdesurvey}
Schouten, K.\BBACOMMA\  \BBA\ Frasincar, F. \BBOP2015\BBCP.
\newblock \BBOQ Survey on aspect-level sentiment analysis\BBCQ\
\newblock {\Bem IEEE Transactions on Knowledge and Data Engineering}, {\Bem
  28\/}(3), 813--830.

\bibitem[\protect\BCAY{Song, Wang, Jiang, Liu,\ \BBA\ Rao}{Song
  et~al.}{2019}]{aen-bert}
Song, Y., Wang, J., Jiang, T., Liu, Z., \BBA\ Rao, Y. \BBOP2019\BBCP.
\newblock
\newblock \BBOQ Attentional encoder network for targeted sentiment
  classification\BBCQ.

\bibitem[\protect\BCAY{Sun, Zhang, Mensah, Mao,\ \BBA\ Liu}{Sun
  et~al.}{2019}]{CDT}
Sun, K., Zhang, R., Mensah, S., Mao, Y., \BBA\ Liu, X. \BBOP2019\BBCP.
\newblock \BBOQ Aspect-level sentiment analysis via convolution over dependency
  tree\BBCQ\
\newblock In {\Bem Proceedings of the 2019 Conference on Empirical Methods in
  Natural Language Processing and the 9th International Joint Conference on
  Natural Language Processing (EMNLP-IJCNLP)}, \BPGS\ 5679--5688, Hong Kong,
  China. Association for Computational Linguistics.

\bibitem[\protect\BCAY{Tang, Qin, Feng,\ \BBA\ Liu}{Tang
  et~al.}{2016a}]{TDLSTM}
Tang, D., Qin, B., Feng, X., \BBA\ Liu, T. \BBOP2016a\BBCP.
\newblock \BBOQ Effective {LSTM}s for target-dependent sentiment
  classification\BBCQ\
\newblock In {\Bem Proceedings of {COLING} 2016, the 26th International
  Conference on Computational Linguistics: Technical Papers}, \BPGS\
  3298--3307, Osaka, Japan. The COLING 2016 Organizing Committee.

\bibitem[\protect\BCAY{Tang, Qin,\ \BBA\ Liu}{Tang et~al.}{2016b}]{DMN}
Tang, D., Qin, B., \BBA\ Liu, T. \BBOP2016b\BBCP.
\newblock \BBOQ Aspect level sentiment classification with deep memory
  network\BBCQ\
\newblock In {\Bem Proceedings of the 2016 Conference on Empirical Methods in
  Natural Language Processing}, \BPGS\ 214--224, Austin, Texas. Association for
  Computational Linguistics.

\bibitem[\protect\BCAY{Tang, Ji, Li,\ \BBA\ Zhou}{Tang et~al.}{2020}]{DGEDT}
Tang, H., Ji, D., Li, C., \BBA\ Zhou, Q. \BBOP2020\BBCP.
\newblock \BBOQ Dependency graph enhanced dual-transformer structure for
  aspect-based sentiment classification\BBCQ\
\newblock In {\Bem Proceedings of the 58th Annual Meeting of the Association
  for Computational Linguistics}, \BPGS\ 6578--6588, Online. Association for
  Computational Linguistics.

\bibitem[\protect\BCAY{Tang, Lu, Su, Ge, Song, Sun,\ \BBA\ Luo}{Tang
  et~al.}{2019}]{songle2019}
Tang, J., Lu, Z., Su, J., Ge, Y., Song, L., Sun, L., \BBA\ Luo, J.
  \BBOP2019\BBCP.
\newblock \BBOQ Progressive self-supervised attention learning for aspect-level
  sentiment analysis\BBCQ\
\newblock In {\Bem Proceedings of the 57th Annual Meeting of the Association
  for Computational Linguistics}, \BPGS\ 557--566, Florence, Italy. Association
  for Computational Linguistics.

\bibitem[\protect\BCAY{Tian, Chen,\ \BBA\ Song}{Tian et~al.}{2021a}]{tgcn}
Tian, Y., Chen, G., \BBA\ Song, Y. \BBOP2021a\BBCP.
\newblock \BBOQ Aspect-based sentiment analysis with type-aware graph
  convolutional networks and layer ensemble\BBCQ\
\newblock In {\Bem Proceedings of the 2021 Conference of the North American
  Chapter of the Association for Computational Linguistics: Human Language
  Technologies (NAACL)}, \BPGS\ 2910--2922.

\bibitem[\protect\BCAY{Tian, Chen,\ \BBA\ Song}{Tian et~al.}{2021b}]{kvmn-eacl}
Tian, Y., Chen, G., \BBA\ Song, Y. \BBOP2021b\BBCP.
\newblock \BBOQ Enhancing aspect-level sentiment analysis with word
  dependencies\BBCQ\
\newblock In {\Bem Proceedings of the 16th Conference of the European Chapter
  of the Association for Computational Linguistics: Main Volume}, \BPGS\
  3726--3739, Online. Association for Computational Linguistics.

\bibitem[\protect\BCAY{Vaswani, Shazeer, Parmar, Uszkoreit, Jones, Gomez,
  Kaiser,\ \BBA\ Polosukhin}{Vaswani et~al.}{2017}]{transformer}
Vaswani, A., Shazeer, N., Parmar, N., Uszkoreit, J., Jones, L., Gomez, A.~N.,
  Kaiser, L., \BBA\ Polosukhin, I. \BBOP2017\BBCP.
\newblock \BBOQ Attention is all you need\BBCQ\
\newblock In {\Bem Advances in Neural Information Processing Systems 30: Annual
  Conference on Neural Information Processing Systems 2017, December 4-9, 2017,
  Long Beach, CA, {USA}}, \BPGS\ 5998--6008.

\bibitem[\protect\BCAY{Velickovic, Cucurull, Casanova, Romero, Li{\`{o}},\
  \BBA\ Bengio}{Velickovic et~al.}{2018}]{gat}
Velickovic, P., Cucurull, G., Casanova, A., Romero, A., Li{\`{o}}, P., \BBA\
  Bengio, Y. \BBOP2018\BBCP.
\newblock \BBOQ Graph attention networks\BBCQ\
\newblock In {\Bem 6th International Conference on Learning Representations,
  {ICLR} 2018, Vancouver, BC, Canada, April 30 - May 3, 2018, Conference Track
  Proceedings}. OpenReview.net.

\bibitem[\protect\BCAY{Wang, Shen, Yang, Quan,\ \BBA\ Wang}{Wang
  et~al.}{2020}]{RGAT}
Wang, K., Shen, W., Yang, Y., Quan, X., \BBA\ Wang, R. \BBOP2020\BBCP.
\newblock \BBOQ Relational graph attention network for aspect-based sentiment
  analysis\BBCQ\
\newblock In {\Bem Proceedings of the 58th Annual Meeting of the Association
  for Computational Linguistics}, \BPGS\ 3229--3238, Online. Association for
  Computational Linguistics.

\bibitem[\protect\BCAY{Wang, Mazumder, Liu, Zhou,\ \BBA\ Chang}{Wang
  et~al.}{2018}]{tsmn}
Wang, S., Mazumder, S., Liu, B., Zhou, M., \BBA\ Chang, Y. \BBOP2018\BBCP.
\newblock \BBOQ Target-sensitive memory networks for aspect sentiment
  classification\BBCQ\
\newblock In {\Bem Proceedings of the 56th Annual Meeting of the Association
  for Computational Linguistics (Volume 1: Long Papers)}, \BPGS\ 957--967,
  Melbourne, Australia. Association for Computational Linguistics.

\bibitem[\protect\BCAY{Wang, Huang, Zhu,\ \BBA\ Zhao}{Wang et~al.}{2016}]{ATAE}
Wang, Y., Huang, M., Zhu, X., \BBA\ Zhao, L. \BBOP2016\BBCP.
\newblock \BBOQ Attention-based {LSTM} for aspect-level sentiment
  classification\BBCQ\
\newblock In {\Bem Proceedings of the 2016 Conference on Empirical Methods in
  Natural Language Processing}, \BPGS\ 606--615, Austin, Texas.

\bibitem[\protect\BCAY{Xing, Liao, Song, Wang, Zhang, Wang,\ \BBA\ Huang}{Xing
  et~al.}{2019}]{AA-LSTM}
Xing, B., Liao, L., Song, D., Wang, J., Zhang, F., Wang, Z., \BBA\ Huang, H.
  \BBOP2019\BBCP.
\newblock \BBOQ Earlier attention? aspect-aware {LSTM} for aspect-based
  sentiment analysis\BBCQ\
\newblock In Kraus, S.\BED, {\Bem Proceedings of the Twenty-Eighth
  International Joint Conference on Artificial Intelligence, {IJCAI} 2019,
  Macao, China, August 10-16, 2019}, \BPGS\ 5313--5319. ijcai.org.

\bibitem[\protect\BCAY{Xue\ \BBA\ Li}{Xue\ \BBA\ Li}{2018}]{gcae}
Xue, W.\BBACOMMA\  \BBA\ Li, T. \BBOP2018\BBCP.
\newblock \BBOQ Aspect based sentiment analysis with gated convolutional
  networks\BBCQ\
\newblock In {\Bem Proceedings of the 56th Annual Meeting of the Association
  for Computational Linguistics}, \BPGS\ 2514--2523, Melbourne, Australia.
  Association for Computational Linguistics.

\bibitem[\protect\BCAY{Zhang, Li, Xu, Leung, Chen,\ \BBA\ Ye}{Zhang
  et~al.}{2020}]{kgcap}
Zhang, B., Li, X., Xu, X., Leung, K.-C., Chen, Z., \BBA\ Ye, Y. \BBOP2020\BBCP.
\newblock \BBOQ Knowledge guided capsule attention network for aspect-based
  sentiment analysis.\BBCQ\
\newblock {\Bem IEEE ACM Trans. Audio Speech Lang. Process.}, {\Bem 28},
  2538--2551.

\bibitem[\protect\BCAY{Zhang, Li,\ \BBA\ Song}{Zhang et~al.}{2019}]{asgcn}
Zhang, C., Li, Q., \BBA\ Song, D. \BBOP2019\BBCP.
\newblock \BBOQ Aspect-based sentiment classification with aspect-specific
  graph convolutional networks\BBCQ\
\newblock In {\Bem Proceedings of the 2019 Conference on Empirical Methods in
  Natural Language Processing and the 9th International Joint Conference on
  Natural Language Processing (EMNLP-IJCNLP)}, \BPGS\ 4568--4578, Hong Kong,
  China. Association for Computational Linguistics.

\bibitem[\protect\BCAY{Zhang\ \BBA\ Qian}{Zhang\ \BBA\ Qian}{2020}]{bigcn}
Zhang, M.\BBACOMMA\  \BBA\ Qian, T. \BBOP2020\BBCP.
\newblock \BBOQ Convolution over hierarchical syntactic and lexical graphs for
  aspect level sentiment analysis\BBCQ\
\newblock In {\Bem Proceedings of the 2020 Conference on Empirical Methods in
  Natural Language Processing (EMNLP)}, \BPGS\ 3540--3549, Online. Association
  for Computational Linguistics.

\end{thebibliography}
